\documentclass[journal]{IEEEtran}
\usepackage{graphicx}
\usepackage{color}
\usepackage{amsmath}
\usepackage{amssymb}
\usepackage{algorithm}
\usepackage{algorithmic}
\usepackage{multirow}
\usepackage{times}
\usepackage{epsfig}

\graphicspath{{Img/}}

\ifCLASSINFOpdf
\else
\fi
\begin{document}
\title{Large Margin Learning in Set to Set Similarity Comparison for Person Re-identification}
\author{Sanping~Zhou,
        Jinjun~Wang,
        Rui~Shi,
        Qiqi~Hou,
        Yihong~Gong,
        Nanning~Zheng
\thanks{Sanping~Zhou, Jinjun~Wang, Rui~Shi, Qiqi~Hou, Yihong~Gong and Nanning~Zheng are all with Institute of Artificial Intelligence and Robotics, Xi'an Jiaotong University, Xi'an, Shaanxi, China.}
\thanks{Corresponding author: Jinjun~Wang. Email: jinjun@mail.xjtu.edu.cn.}}

\markboth{IEEE Transactions on Multimedia}%
{Shell \MakeLowercase{\textit{et al.}}: Bare Demo of IEEEtran.cls for Journals}
\maketitle

\begin{abstract}
Person re-identification~(Re-ID) aims at matching images of the same person across disjoint camera views, which is a challenging problem in multimedia analysis, multimedia editing and content-based media retrieval communities. The major challenge lies in how to preserve similarity of the same person across video footages with large appearance variations, while discriminating different individuals. To address this problem, conventional methods usually consider the pairwise similarity between persons by only measuring the point to point~(P2P) distance. In this paper, we propose to use deep learning technique to model a novel set to set~(S2S) distance, in which the underline objective focuses on preserving the compactness of intra-class samples for each camera view, while maximizing the margin between the intra-class set and inter-class set. The S2S distance metric is consisted of three terms, namely the class-identity term, the relative distance term and the regularization term. The class-identity term keeps the intra-class samples within each camera view gathering together, the relative distance term maximizes the distance between the intra-class class set and inter-class set across different camera views, and the regularization term smoothness the parameters of deep convolutional neural network~(CNN). As a result, the final learned deep model can effectively find out the matched target to the probe object among various candidates in the video gallery by learning discriminative and stable feature representations. Using the CUHK01, CUHK03, PRID2011 and Market1501 benchmark datasets, we extensively conducted comparative evaluations to demonstrate the advantages of our method over the state-of-the-art approaches.
\end{abstract}


\IEEEpeerreviewmaketitle

\section{Introduction}

\begin{figure}[!htb]
\footnotesize
\centering
\includegraphics[height = 6.0cm, width = 8.0cm]{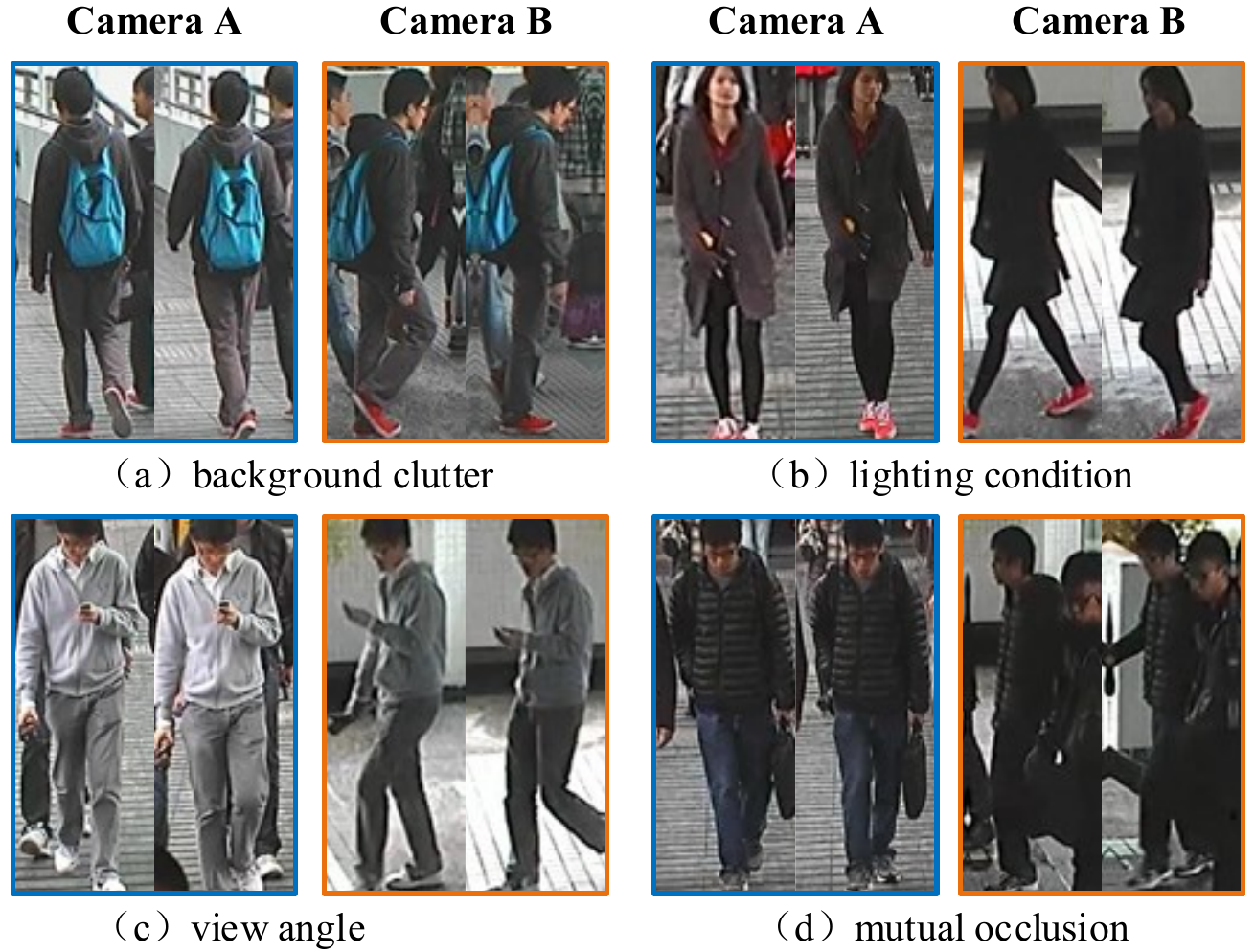}\\
\caption{The challenges to person re-identification in public space, where the blue bounding boxes denote the training samples in Camera A and the orange bounding boxes denote the training samples in Camera B.}
\label{fig_1}
\end{figure}

\IEEEPARstart{G}{iven} one single shot or multiple shots of an object, person re-identification~(Re-ID) aims at matching the stated person among various candidates from a set of disjoint camera views~\cite{Wang_Hu_Liang:2016,Ye_Liang_Yu:2016}. Person Re-ID is a key technology to many advanced multimedia applications, such as person association~\cite{Morris_Trivedi:2008}, multi-target tracking~\cite{Zhang_Wang_Wang:2015} and behavior analysis~\cite{Hu_Tan_Wang:2004}. In theses tasks, the appearance of an object usually possesses large variations due to background clutter, lighting condition, view angle and mutual occlusion~\cite{Sunderrajan_Manjunath:2016}, as shown in Fig.~\ref{fig_1}. These challenges have made person Re-ID a very difficult problem, which has attracted a lot of attention from researchers in computer vision. The key to improve the identification performance is to learn both discriminative and stable feature representation to describe the appearance of person, in which the difference between intra/inter-class distance can be maximized.

Existing person Re-ID works can be broadly grouped into the following two lines: 1) developing robust feature representation to handle the variations in persons' appearance, and 2) designing discriminative distance metrics to measure the similarity between intra/inter-class image distance. In the first category, different cues have been adopted to extract stable and discriminative features, and representative works including the Local Binary Pattern~(LBP)~\cite{Xiong_Gou_Camps:2014}, Ensemble of Local Feature~(ELF)~\cite{Gray_Tao:2008} and Local Maximal Occurrence~(LOMO)~\cite{Zhao_Ouyang_Wang:2014}; In the second category, labeled images are used to train a distance metric, in which the intra-class distance is minimized while the inter-class distance is maximized. Typical metric learning methods include the Locally Adaptive Decision Function~(LADF)~\cite{Li_Chang_Liang:2013}, Large Margin Nearest Neighbor~(LMNN)~\cite{Weinberger_Blitzer_Saul:2005} and Information Theoretic Metric Learning~(ITML)~\cite{Davis_Kulis_Jain:2007}. Since both lines of works regard the feature extraction and metric learning processes as two disjoint steps, their performance is limited.

Recently, the deep convolutional neural network~(CNN) based methods are becoming popular in solving the person Re-ID problem~\cite{Ahmed_Jones_Marks:2015,Ding_Lin_Wang:2015,Zhang_Lin_Zhang:2015,Zhou_Wang_Hou:2016}. Although these methods usually consist of two major components analogue to that of the traditional person Re-ID literature, i.e., a deep CNN to extract the appearance feature and a similarity measure to enforce the learned feature to be similar/dissimilar for intra/inter-class samples, by incorporating the feature extraction step and the metric learning step into one integrated framework, these deep CNN based method exhibits significantly improved performance for person Re-ID, where the extracted features are robust to the appearance variations across different camera views.

Despite the progress achieved by these deep learning based methods, it is observed that the lacking of labeled training data usually limits their generalization ability in practical person Re-ID applications. The situation has motivated us to focus on the following two aspects to improve the generalization ability of a person Re-ID system: 1) the CNN should not be too deep, and 2) the loss function should incorporate additional prior knowledge to be more efficient at training. In this paper, we propose a novel set to set~(S2S) distance metric to supervise the trained CNN to learn more discriminative and robust appearance feature based on the same amount of training data. Specifically, we first build a part-based CNN to extract the image feature in which different body parts are separately modeled in lower convolutional layers and then fused in higher fully connected layers. The resulting feature representations are then fed into the S2S loss layer which focuses on preserving the compactness of intra-class samples for each camera view, while maximizing the margin between the intra-class set and inter-class set. In this way, our method naturally incorporates the feature extraction and metric learning into a joint framework. By considering the set to set similarity, the proposed model can extract features that is robust to the appearance variations as evidenced by the improved person Re-ID performance for several public benchmark datasets.

In summary, the main components of our method includes:

\begin{itemize}
  \item A novel part-based CNN to extract discriminative and stable features for the body appearance. The CNN constitutes of a global sub-network, a local sub-network, and a fusion sub-network. Global and local feature representations are firstly extracted in the global sub-network and local sub-network respectively, and then fused in the fusion sub-network. We also make use of deep residual neural network~\cite{He_Zhang_Ren:2016} to construct our part-based CNN.
  \item A novel S2S loss layer for similarity comparison. It preserves the compactness of intra-class samples, and keeps a large margin between the intra-class set and inter-class set of training samples by maximizing the relative distance between samples from the margins of both the intra-class set and inter-class set.
  \item Extensive comparative experiments to evaluate various aspects of our method on four benchmark datasets, including the CUHK01, CUHK03, PRID2011, and Market1501. The final evaluation results show that our method outperforms the state-of-the-art person Re-ID methods by a large margin.
\end{itemize}

The rest of our paper is organized as follows. In Section~\ref{sec_rel}, we briefly review the related works. Section~\ref{sec_alog} introduces our deep feature learning and fusion network, and the S2S distance metric, followed by a discussion of the learning algorithm in Section~\ref{sec_alog}. Experimental results and parameter analysis are presented in Section~\ref{sec_exp}. And conclusion comes in Section~\ref{sec_con}.

\section{Related Work}
\label{sec_rel}

The past few years of works closely related to our way in solving the person Re-ID problem can be organized into four classes, namely the Feature Learning based Method, the Metric Learning based Method, the Deep Learning based Method and the Set-Based Classification Method. The following paragraphs respectively introduce these works.

{\bf Feature Learning based Method} The line of works mainly focus on developing robust feature representations which are invariant to view angles, lighting conditions, body poses and background variations. For examples, Zhao et al.~\cite{Zhao_Ouyang_Wang:2014} learned a mid-level filter from patch cluster to achieve cross view invariance. In~\cite{Liao_Hu_Zhu:2015}, Liao et al. constructed a feature descriptor which analyzed the horizontal occurrence of local features and maximized the occurrence to make a robust representation against viewpoint changes. Ma et al.~\cite{Ma_Su_Jurie:2012} presented the person image via covariance descriptor which was robust to illumination changes and background variations. In~\cite{Farenzena_Bazzani_Perina:2010}, Farenzena et al. augmented maximally stable color regions with histograms for person representation. Zhao et al.~\cite{Zhao_Ouyang_Wang:2013} learned the distinct salience feature to distinguish the matched person from others. In~\cite{Cheng_Cristani_Stoppa:2011}, Chen et al. employed a pre-learned pictorial structure model to localize the body parts more accurately. Wu et al.~\cite{Wu_Li_Radke:2015} introduced a viewpoint-invariant descriptor, which took the viewpoint of the human into account by using what they called a pose prior learned from the training data. In~\cite{Koestinger_Hirzer_Wohlhart:2012}, Kviatkovsky et al. investigated the intra-distribution structure of color descriptor, which was invariant under certain illumination changes. Li et al.~\cite{Li_Wang:2013} matched person images observed in different camera views with a complex cross-view transformation and applied it to person Re-ID. These methods aim to improve the performance of person Re-ID by developing a fixed robust feature descriptor, however adaptive feature learning is not addressed in these methods.

{\bf Metric Learning based Method} The metric learning based methods aim to find a mapping function from the feature space to distance space, in which feature vectors from the same person are closer than those from different ones~\cite{Zheng_Gong_Xiang:2011}. For example, Zheng et al.~\cite{Zheng_Gong_Xiang:2013} proposed a relative distance learning method from a probabilistic prospective. In~\cite{Mignon_Jurie:2012}, Mignon et al. learned a distance metric from sparse pairwise similarity constraints. Pedagadi et al.~\cite{Pedagadi_Orwell_Velastin:2013} utilized the Local Fisher Discriminant Analysis~(LFDA) to map high dimensional features into a more discriminative low dimensional space. In ~\cite{Xiong_Gou_Camps:2014}, Xiong et al. further extended the LFDA and several other metric learning methods by using the kernel tricks and different regularizers. Nguyen et al.~\cite{Nguyen_Bai:2011} measured the similarity of face pairs through cosine similarity, which was closely related to the inner product similarity. In~\cite{Loy_Liu_Gong:2013}, Loy et al. casted the person Re-ID problem as an image retrieval task by considering the listwise similarity. Chen et al.~\cite{Chen_Yuan_Hua:2015} proposed a kernel based metric learning method to explore the nonlinearity relationship of samples in the feature space. In~\cite{Hirzer_Roth:2012}, Hirzer et al. learned a discriminative metric by using relaxed pairwise constraints. Prosser et al. developed~\cite{Prosser_Zheng_Gong:2010} a ranking model using support vector machine. These methods learn a specific distance metric mainly based on feature representation extracted by several fixed feature descriptors, which may influence the performance of metric learning.

{\bf Deep Learning based Method} The deep learning based methods aim to incorporate feature extraction and metric learning into an integrated framework, in which adaptive feature representations can be learned under the supervision of a certain similarity metric. For example, Li et al.~\cite{Li_Zhao_Xiao:2014} proposed a novel Filter Pairing Neural Network~(FPNN) to model body part displacements by using the patch matching layers to match the filter responses of local patches across camera views. In~\cite{Ahmed_Jones_Marks:2015}, Ahmed et al. proposed an improved deep neural network which took the pairwise images as inputs, and output a similarity value indicating whether the two input images depict the same person or not. Xiao et al.~\cite{Xiao_Li_Ouyang:2016} proposed a domain guided dropout algorithm to improve the performance of deep CNN to extract robust feature representation for person Re-ID. In~\cite{Yi_Lei_Liao:2014}, Yi et al. constructed a siamese neural network to learn the pairwise similarity, and used body parts to train the model. Ding et al.~\cite{Ding_Lin_Wang:2015} applied the triplet loss to train the deep neural network for person Re-ID. In~\cite{Wang_Zuo_Lin:2016}, Wang et al. proposed a unified triplet and siamese deep architecture which could jointly extract single-image and cross-image feature representations. Zhao et al.~\cite{Zhao_Ouyang_Wang:2014} proposed a local patch matching method which used a learned the mid-level filters to get the local discriminative features for person Re-ID. In~\cite{Zhang_Lin_Zhang:2015}, Zhang et al. incorporated the deep hash learning into a triplet formulation for person Re-ID. Chen et al.~\cite{Chen_Guo_Lai:2016} proposed a unified deep ranking model where feature extraction and metric learning were handled into a joint deep learning framework. In these methods, feature extraction and metric learning are incorporated into a joint framework mainly based on the general deep CNN, such as AlexNet~\cite{Krizhevsky_Sutskever_Hinton:2012}, without using an effective part strategy, which may be inappropriate for person Re-ID.

{\bf Set-Based Classification Method} Finally, the set-based classification methods aim at exploring the S2S relationship, so as to well handle the intra-class or inter-class appearance variations. Different from the conventional point-based methods, which model the intra-class or inter-class appearance variations by only considering the point to point~(P2P) distance. For example, Zhou et al.~\cite{Zhou_Wang_Wang:2017} proposed a point to set distance metric, which measures the similarity between an anchor point and the corresponding positive and negative set for person Re-ID. In~\cite{Zhu_Zhang_Zuo:2013}, Zhu et al. extended the P2P distance metric to S2S distance metric and incorporated it into the standard support vector machine. Zhu et al.~\cite{Zhu_Zuo_Zhang:2014} proposed a novel image set based collaborative representation and classification method by modeling the query set as convex or regularized hull for face recognition. In~\cite{Wang_Shan_Chen:2008}, Wang et al. measured the manifold to manifold distance and applied it in face recognition based on image set. Wu et al.~\cite{Wu_Minoh_Mukunoki:2012} proposed a set-based discriminative ranking method to deal with the person Re-ID and face recognition problem. In~\cite{Yang_Wang_Liu:2017}, Yang et al. regularized the nearest points for image set based face recognition. Lu et al.~\cite{Lu_Li_Fang:2016} regarded the spectrally similar pixels within each homogeneous as one set of samples, and proposed a novel set-based spectral-spatial classification method for hyper spectral images. In~\cite{Lu_Wang_Deng:2015}, Lu et al. proposed a multi-manifold deep metric learning method for image set classification, which could recognize an object from a set of image instances with varying viewpoints or illuminations. Mian et al.~\cite{Mian_Hu_Hartley:2013} proposed a self-regularized nonnegative coding to define between set distance by measuring between the nearest set points for robust face recognition. These methods usually define a kind of set-based distance in the traditional metric learning methods, and we do not see its wide combination with the popular deep learning methods.

\begin{figure*}[!htb]
\centering
    \begin{tabular}{cc}
        \hspace{-0.4cm}
        \includegraphics[height = 7.2cm, width = 18.0cm]{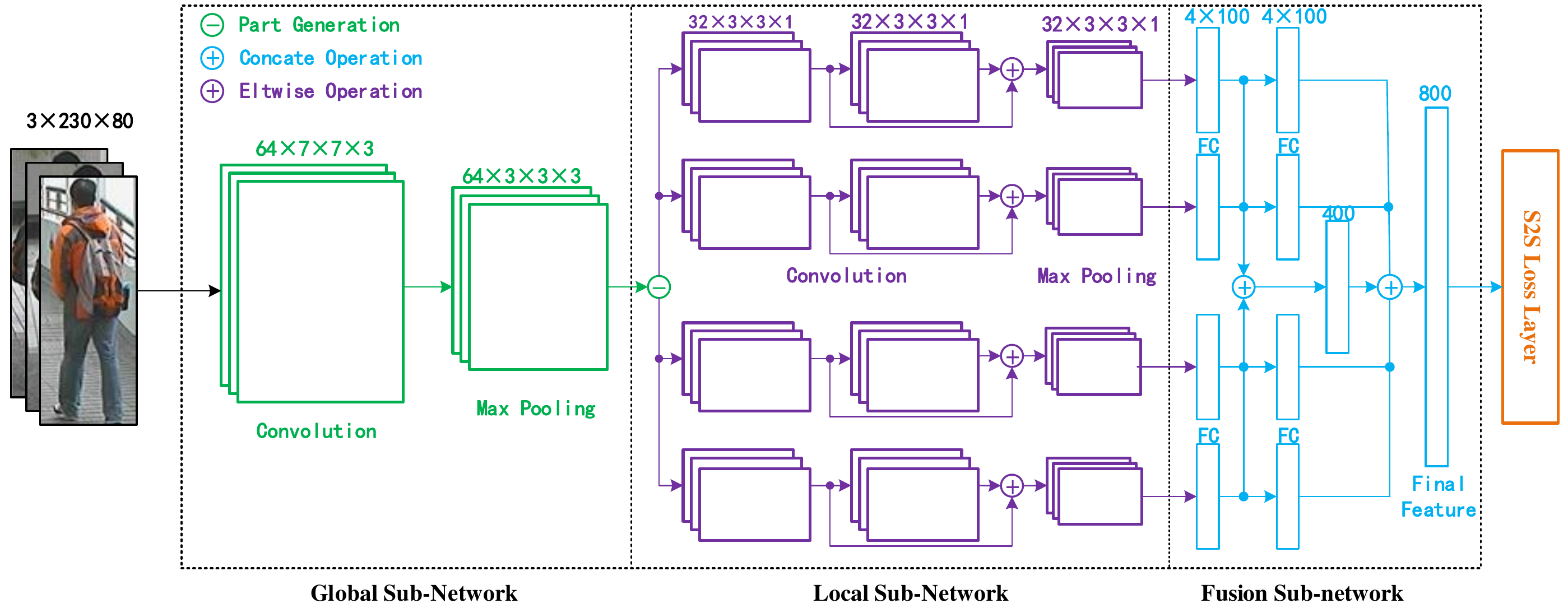}
    \end{tabular}
    \caption{The deep feature extraction and fusion neural network. This architecture is comprised of three sub-networks: the global sub-network, the local sub-network and the fusion sub-network. The global sub-network and local sub-network use the convolutional layers and max-pooling layers to extract global and local feature representations of an object. The fusion sub-network uses fully connected layers to generate a final combined feature representations from each body parts as well as from the entire body. Finally, the resulting features are fed into the S2S loss layer for set to set similarity comparison}
    \label{fig_2}
\end{figure*}
\section{The Proposed Deep Neural Network and Set to Set Distance Metric}
\label{sec_alog}

\subsection{The Deep Neural Network}
In order to incorporate feature learning and fusion into an end-to-end framework, we introduce a novel part-based deep CNN to extract discriminative and robust feature representation from an object, as shown in Fig.~\ref{fig_2}. Given a set of training samples, the task is to simultaneously compute the compactness of intra-class samples (i.e., those from the same camera view) and maximize the margin between the intra-class set and inter-class set across different camera views.

{\bf Global sub-network} The first part of our neural network is a global sub-network, which consists of a convolutional layer and a max pooling layer. They are designed to bridge the low-level features of the raw input images, so as to provide multi-level feature representations to be discriminately learned in the following local sub-network. The input images are in size of $230\times80\times 3$, and are firstly passed through $64$ learned filters of size $7\times7\times3$. Then, the resulting feature maps are passed through a max pooling kernel of size $3\times3$ with stride $3$. Finally, these feature maps are passed through a rectified linear unit~(ReLU).

{\bf Local sub-network} The second part of our neural network is a local sub-network, which is consisted of four teams of convolutional layers and max pooling layers. We firstly divide the input feature maps into four equal horizontal patches across the height channel, which introduces $4\times64$ local feature maps of different body parts. Then, we pass each local feature maps through two convolutional layers, and both of them have $32$ learned filters of size $3\times3\times1$. Afterwards, the outputs of the first local convolutional neural network are summarized with the outputs of the second local convolutional neural network using the eltwise operation. The resulting feature maps are further passed through max pooling kernels of size $3\times3$ with stride 1. Finally, we add a rectified linear unit~(ReLU) after each of the max pooling layers. In order to learn the feature representations of different body parts discriminatively, we do not share the parameters among the four teams of convolutional neural layers.

{\bf Fusion sub-network} The third part of our neural network is a fusion sub-network, which is consisted of four teams of fully connected layers. Firstly, the local feature maps of different body parts are discriminatively learned by concatenating two fully connected layers in each team. The dimension of these fully connected layers is $100$ and a rectified linear unit~(ReLU) is added between the two fully connected layers. Then, the discriminatively learned local feature representations of the first four fully connected layers are concated to be summarized by adding another fully connected layers, whose dimension is 400. Finally, the resulting feature representation is further concated with the outputs of the second four fully connected layers to generate final 800 dimensional feature representations. Similarly, we do not share the parameters among the four fully connected layers to keep the discriminative of feature representations of different body parts.
\subsection{The S2S Distance Metric}
To begin with, let $\mathbf{X} = \{\mathbf{X}_i\}_{i=1}^N$ be the input training set samples, where $N$ denotes the number of different persons and $\mathbf{X}_i=\{\mathbf{x}_{i,j}|(\mathbf{x}_{i,j}^A, \mathbf{x}_{i,j}^B)\}_{j=1}^M$ represents the $M$ training samples of each person from camera view $A$ and camera view $B$~\footnote{Although the proposed algorithm can handle multiple camera views, we present our discussion within two camera views situation and assume that there are same number of images for every person under both of the two camera views for simplicity.}. The goal of our deep neural network is to learn filter weights and biases that minimizes the ranking error from the output layer. A recursive function for an $K$-layer deep model can be formulated as follows:
\begin{equation}
\label{eq_1}
    \begin{aligned}
        \mathbf{X}_i^{(k)} &= \Psi(\mathbf{W}^{(k)}*\mathbf{X}_i^{(k-1)} + \mathbf{b}^{(k)}),\\
        i = 1, 2, \cdots&, N; k = 1, 2, \cdots, K; \mathbf{X}_i{(0)} = \mathbf{X}_i.
    \end{aligned}
\end{equation}
where $\mathbf{W}^{(k)}$ denotes the filter weights of the $k^{th}$ layer, $\mathbf{b}^{(k)}$ refers to the corresponding biases, $*$ denotes the convolution operation, $\Psi(\cdot)$ is an element-wise non-linear activation function such as ReLU, and $\mathbf{X}_i^{(k)}$ represents the feature maps generated at layer $k$ for sample $\mathbf{X}_i$. For similarity, we simplify the parameters of the neural network as a whole and define $\mathbf{W} = \{\mathbf{W}^{(1)}, \cdots, \mathbf{W}^{(K)}\}$ and $\mathbf{b} = \{\mathbf{b}^{(1)}, \cdots, \mathbf{b}^{(K)}\}$.

\begin{figure*}[!htb]
\centering
    \begin{tabular}{c}
        \includegraphics[height = 5.5cm, width = 17.5cm]{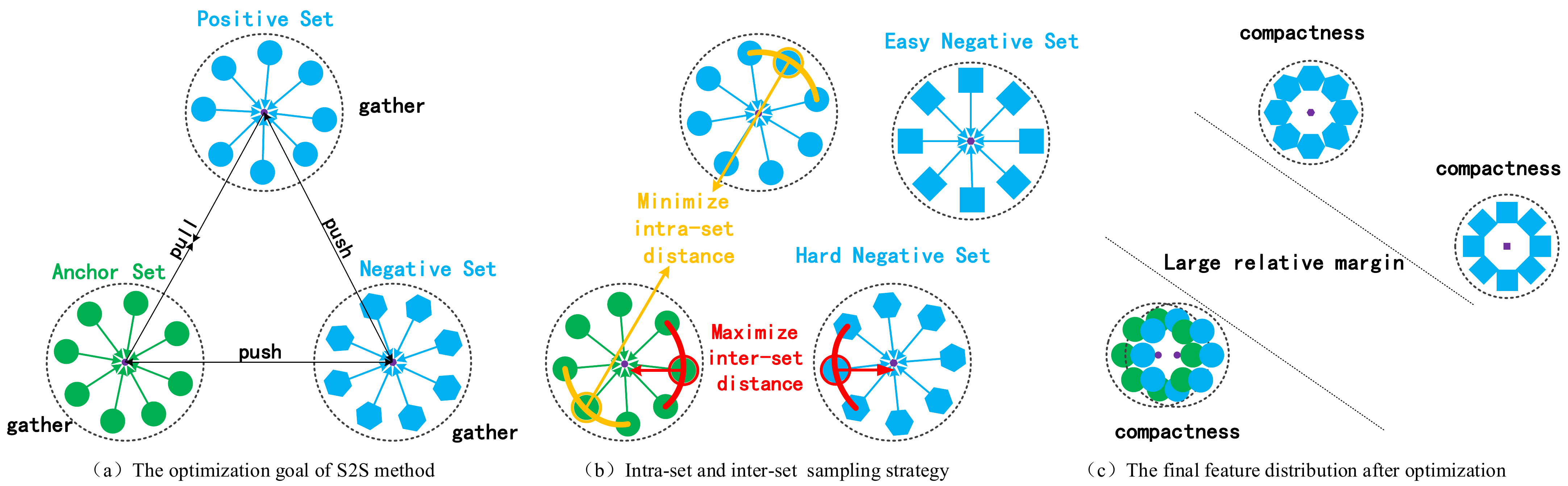}
    \end{tabular}
    \caption{Illustration of our S2S metric. Samples from different camera views are denoted in different colors and different identities are denoted by different shapes. Specially, (a) shows the optimization goal of our method, namely maintaining the compactness of intra-class samples under each camera view and maximizing a large margin between the intra-class set and inter-class set across different camera views; (b) defines the set to set distance in our S2S distance metric, including how to choose the marginal samples and the hard negative set; (c) represents the final feature distributions of the training samples in feature space.}
    \label{fig_3}
\end{figure*}

The basic idea of our S2S distance metric is shown in Fig.~\ref{fig_3}, where we supervise the learning process of the deep CNN by keeping the compactness of intra-class samples under each camera view and maximizing a large relative margin between the intra-class set and inter-class set. The S2S metric is consisted of three terms, namely the class-identity term, the relative distance term and the regularization term, which can be formulated as follows:
\begin{equation}
\label{eq_2}
     \mathrm{L} = \alpha \mathrm{L_C}(\mathbf{X},\mathbf{W},\mathbf{b}) + \mathrm{L_S}(\mathbf{X},\mathbf{W},\mathbf{b}) + \beta \mathrm{R}(\mathbf{W},\mathbf{b}),
\end{equation}
where $\mathrm{L_C}(\cdot)$ is the class-identity term, $\mathrm{L_S}(\cdot)$ denotes the relative distance term, $\mathrm{R}(\cdot)$ represents the regularization term, and $\alpha, \beta$ are two constant weight parameters. In the training process, the class-identity term keeps the compactness of intra-class training samples under each camera view, the relative distance term maximizes the distance between the intra-class set and inter-class set, and the regularization term smoothes the parameters of the deep CNN.

{\bf The class-identity term} In order to strengthen the compactness of the intra-class samples under each camera view, we assume that they should be gathered together in the training process. Therefore, the hinge-like loss of the class-identity term can be formulated as follows:
\begin{equation}
\label{eq_3}
    \mathrm{L_C} = \frac{1}{\mathcal{Z}_c}\sum\limits_{i = 1}^N \sum\limits_{j = 1}^M  \max\{\|{\mathbf{c}_i} - \mathbf{x}_{i,j}\|_2^2 - \mathcal{M}_c,0\},
\end{equation}
where $\mathcal{Z}_c$ is a normalization factor, $\mathbf{c}_i=\frac{1}{M}\sum\nolimits_{j = 1}^M{\mathbf{x}_{i,j}}$ denotes the $i^{th}$ class center, and $\mathcal{M}_c$ is a constant margin parameter. As a result, the feature representations of the intra-class training samples under each camera view should be gathered together, and therefore the compactness of set samples is held.

\begin{figure}[!htb]
\footnotesize
\centering
\includegraphics[height = 4.2cm, width = 8.5cm]{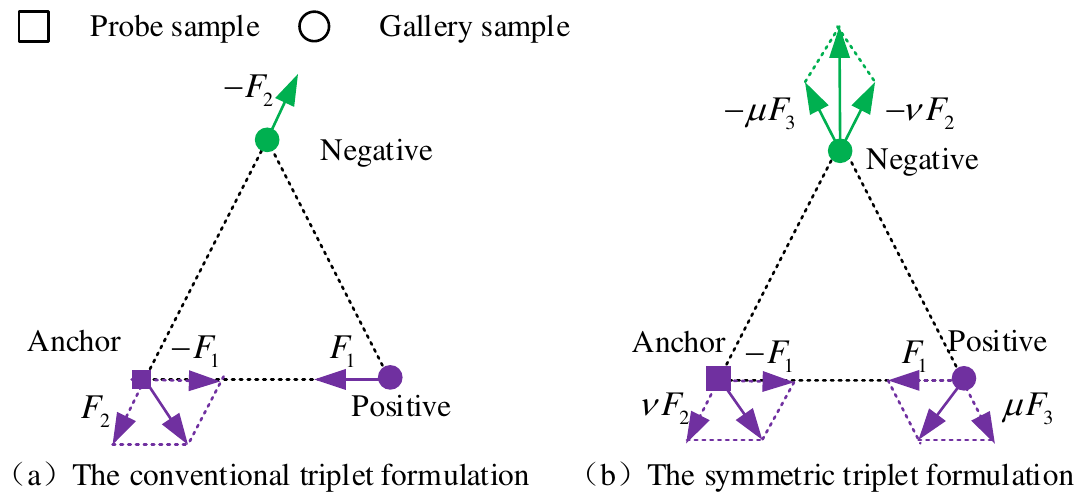}\\
\caption{The comparison of the conventional triplet formulation and the symmetric triplet formulation. Specially, (a) and (b) shows the gradient flow of each training sample in a triplet unit, in which $\mathbf{F}_1 = 2(\mathbf{x}_{i,a}^A-\mathbf{x}_{i,p}^B)$, $\mathbf{F}_2 = 2(\mathbf{x}_{i,a}^A-\mathbf{x}_{i,n}^B)$ and $\mathbf{F}_3 = 2(\mathbf{x}_{i,p}^B-\mathbf{x}_{i,n}^B)$ denote the three basic components.}
\label{fig_4}
\end{figure}

{\bf The relative distance term} The goal of our relative distance term is to maximize the relative distance between the intra-class set and inter-class set. Considering the fact that there is no unambiguous definition of the S2S distance, we formulate the relative distance term as follows:

\begin{equation}
\label{eq_4}
    \mathrm{L_S} = \mathrm{L_T}(\mathbf{X},\mathbf{W},\mathbf{b}) + \lambda \mathrm{L_P}(\mathbf{X},\mathbf{W},\mathbf{b}),
\end{equation}
where $\mathrm{L_T}(\cdot)$ denotes the triplet distance term, which is designed to keep the stability of the deep CNN by randomly choosing samples from the image sets in a symmetric triplet formulation; $\mathrm{L_P}(\cdot)$ represents the pairwise distance term, which is used to boost the ranking performance of the deep CNN by adaptively choosing the training samples from margins of the image set in a pairwise formulation; and $\lambda$ is a constant weight parameter.

{\bf Definition}-{\textsl Symmetric Triplet:}\footnote{As shown in Fig.~\ref{fig_4}, the symmetric triplet formulation outperforms the conventional one by improving the gradient back-propagations of samples in each triplet unit.} Given a set of triplet training samples $\{\mathbf{x}_{i,a}^A, \mathbf{x}_{i,p}^B, \mathbf{x}_{i,n}^B\}_{i=1}^N$, in which $\{\mathbf{x}_{i,a}^A,\mathbf{x}_{i,p}^B\}$ is the positive pair and $\{\mathbf{x}_{i,a}^A,\mathbf{x}_{i,n}^B\}$ denotes the negative pair. The conventional triplet formulation maximizes a large relative margin $\|\mathbf{x}_{i,a}^A - \mathbf{x}_{i,n}^B\|_2^2 - \|\mathbf{x}_{i,a}^A - \mathbf{x}_{i,p}^B\|_2^2 \geq M$ by using the loss $ \mathrm{L} = \sum \nolimits_{i=1}^N \max\{M + \|\mathbf{x}_{i,a}^A-\mathbf{x}_{i,p}^B\|_2^2 - \|\mathbf{x}_{i,a}^A-\mathbf{x}_{i,n}^B\|_2^2,0\}$. In our symmetric triplet formulation, we satisfy the above constraint by using the loss $\mathrm{L} = \sum \nolimits_{i=1}^N \max\{M + \|\mathbf{x}_{i,a}^A - \mathbf{x}_{i,p}^B\|_2^2 - [\mu \|\mathbf{x}_{i,a}^A - \mathbf{x}_{i,n}^B\|_2^2 + \nu \|\mathbf{x}_{i,p}^B - \mathbf{x}_{i,n}^B\|_2^2],0\}$, where the first term denotes the intra-class distance, the second term and the third term are weighted to represent the inter-class distance, and $\mu,\nu$ are two adaptive weights.

Considering the fact that the optimization of these marginal samples by only using the pairwise term can easily lead the network fall into a plateau, we propose a triplet distance term which is built to avoid such kind of over-fitting problem by randomly sampling the anchor, the positive and the negative samples in the training image sets. It is formulated in the symmetric triplet formulation as follows:

\begin{equation}
\label{eq_5}
    \mathrm{L_T} = \frac{1}{\mathcal{Z}_t} \hspace{-0.1cm} \sum\limits_{i,j,k = 1}^N \hspace{-0.05cm} \sum\limits_{l,s,r = 1}^M \hspace{-0.1cm} \max\{\mathcal{M}_t -\mathrm{T}(\mathbf{x}_{i,l}^A,\mathbf{x}_{j,s}^B, \mathbf{x}_{k,r}^B),0\},
\end{equation}
where $\mathcal{Z}_t$ is a normalization factor, $\mathcal{M}_t$ is a constant margin parameter, and $\mathrm{T}$ denotes the relative distance between the positive pair and the corresponding negative pair. The relative distance is defined as follows:

\begin{equation}
\label{eq_6}
    \mathrm{T}\hspace{-0.1cm} = \hspace{-.1cm}\mathrm{N}_{i,k}^{l,r}(\mu \|\mathbf{x}_{i,l}^A-\mathbf{x}_{k,r}^B\|_2^2 + \nu \|\mathbf{x}_{j,s}^B - \mathbf{x}_{k,r}^B\|_2^2) - \mathrm{P}_{i,j}^{l,s}\|\mathbf{x}_{i,l}^A - \mathbf{x}_{j,s}^B\|_2^2,
\end{equation}
where $\mu, \nu$ are two adaptively learned weights, and $\mathrm{N}_{i,k}^{l,r}$ and $\mathrm{P}_{i,j}^{l,s}$ denote the negative and positive indicator matrix. Given the anchor $l$ from the $i^{th}$ class in camera view A, $\mathrm{N}_{i,k}^{l,r}$ and $\mathrm{P}_{i,j}^{l,s}$ indicate the randomly chosen negative candidates and positive candidates in camera view B, respectively. They are defined as follows:

\begin{equation}
\label{eq_7}
\begin{aligned}
    &\mathrm{P}_{i,j}^{l,s}= \left\{
    \begin{array}{l}
        1, \hspace{0.1cm} if \hspace{0.1cm}i = j, \hspace{0.05cm} and \hspace{0.1cm} l, s < M;\\
        0, \hspace{0.1cm}else.
    \end{array} \right.\\
    &\mathrm{N}_{i,k}^{l,r}= \left\{
    \begin{array}{l}
        1, \hspace{0.1cm}if \hspace{0.1cm} i \ne k, \hspace{0.05cm} and \hspace{0.1cm} l, r < M;\\
        0, \hspace{0.1cm}else.
    \end{array} \right.
\end{aligned},
\end{equation}
where $\mathrm{P}_{i,j}^{l,s} = 1$ means $\{l,s\}$ is a positive pair, while $\mathrm{N}_{i,k}^{l,r} = 1$ means $\{l,r\}$ is a negative pair; and vice verse. As shown in Fig.~\ref{fig_4}, the gradient back-propagation of $\mathbf{x}_{i,p}^B$ in the conventional triplet term lacks the control from $\mathbf{x}_{i,a}^A$, which makes $\mathbf{x}_{i,p}^B$ lag behind in the vertical direction. Different from the conventional triplet formulation, our symmetric triplet term can well solve this problem by introducing a weighted negative distance to improve the gradient back-propagation of $\mathbf{x}_{i,a}^A, \mathbf{x}_{i,p}^B$ and $\mathbf{x}_{i,n}^B$ in each triplet unit. As a result, the positive distance can be minimized and the negative distance can be maximized simultaneously in our method.

Different from the triplet distance term, the pairwise distance term aims to boost the ranking performance of the deep CNN by taking marginal samples between the intra-class sets and inter-class sets into consideration, as shown in Fig.~\ref{fig_3} (b). It is formulated in the conventional pairwise formulation, and can be defined as follows:

\begin{equation}
\label{eq_8}
    \mathrm{L_P} = \frac{1}{\mathcal{Z}_p} \hspace{-0.1cm} \sum\limits_{i,j = 1}^N \hspace{-0.05cm} \sum\limits_{l,r = 1}^M \max\{\mathcal{C}_p - \mathrm{G}_{i,j}^{l,s}(\mathcal{M}_p - \|\mathbf{x}_{i,l}^A - \mathbf{x}_{j,s}^B\|_2^2),0\},
\end{equation}
where $\mathcal{Z}_p$ is a normalization factor, the parameters $\mathcal{M}_p > \mathcal{C}_p$ are used to define the down-margin and upper-margin. In particular, $\mathcal{M}_p - \mathcal{C}_p$ denotes the down-margin, and $\mathcal{M}_p + \mathcal{C}_p$ represents the up-margin. Given the $i^{th}$ and $j^{th}$ identities, the indicator matrix $\mathrm{G}_{i,j}^{l,s}$ refers to the correspondence of the $s^{th}$ image in camera B to the  $l^{th}$ image in camera A, which is defined as follows:

\begin{equation}
\label{eq_9}
    \mathrm{G}_{i,j}^{l,s}= \left\{
    \begin{array}{l}
        +1, \hspace{0.05cm} if \hspace{0.05cm}j  = \tau_p(i), s = \pi_p(l), \hspace{0.05cm} and \hspace{0.1cm} l,s < M;\\\\
        -1, \hspace{0.05cm} if \hspace{0.05cm} j = \tau_n(i), s = \pi_n(l), \hspace{0.05cm} and \hspace{0.1cm} l,s < M.
    \end{array}, \right.
\end{equation}
where $\mathrm{G}_{i,j}^{l,s} = 1$ means that the $s^{th}$ image of the $j^{th}$ identity in camera view B is referred to the same person to that of the $l^{th}$ image of the $i^{th}$ identity in camera view A, while $\mathrm{G}_{i,j}^{l,s} = -1$ means the opposite. Given the $i^{th}$ class in camera view A, $\tau_p(i)$ and $\tau_n(i)$ respectively denote the farthest positive and the nearest negative candidate sets in camera view B. As shown in Fig.~\ref{fig_3} (b), given the $l^{th}$ marginal sample in camera view A, $\pi_p(l)$ and $\pi_n(l)$ represent the positive and negative candidate images located in between the margin of $j^{th}$ image set in camera view B and $i^{th}$ image set in camera view A, respectively.

{\bf The regularization term} In order to smooth the parameters of the whole neural network, we define the following regularization term,
\begin{equation}
\label{eq_10}
    \mathrm{R} = \sum\limits_{k = 1}^K \| \mathbf{W}^{(k)}\|_F^2 + \| \mathbf{b}^{(k)}\|_2^2,
\end{equation}
where $\|\cdot\|_F^2$ denotes of the Frobenius norm, and $\|\cdot\|_2^2$ represents the Euclid norm.

The S2S distance metric can not only keep the compactness of the intra-class samples under each camera view, but also maximize a large relative margin between the intra-class set and inter-class set across different camera views. Therefore, given a probe image, the learned deep model can easily find out the matched candidate from the gallery set by ranking the distances in the testing process.
\section{Optimization Algorithm}
\label{sec_alog}
We use the gradient back-propagation method to optimize the parameters of the deep CNN, which is carried out in the mini-batch  pattern. Therefore, we need to calculate the gradients of the loss function with respect to the features of the corresponding layers. For simplicity, we consider the parameters in the network as a whole and define $\mathbf{\Omega}^{(k)} = [\mathbf{W}^{(k)}, \mathbf{b}^{(k)}]$, and $\mathbf{\Omega} = \{\mathbf{\Omega}^{(1)}, \dots, \mathbf{\Omega}^{(K)}\}$.

In order to employ the back-propagation algorithm to optimize the network parameters, we compute the partial derivative of the loss function as follows:
\begin{equation}
\label{eq_11}
    \frac{\partial \mathrm{L}}{\partial \mathbf{\Omega}} = \sum\limits_{i = 1}^N \alpha c(\mathbf{X}_i, \mathbf{\Omega}) + s(\mathbf{X}_i, \mathbf{\Omega}) + 2 \beta \sum\limits_{k = 1}^K \mathbf{\Omega}^{(k)},
\end{equation}
where the three terms denote the gradient of the class-identity term, the relative distance term and the regularization term, respectively.

For simplicity, we define $\mathcal{C} = \|{\mathbf{c}_i} - \mathbf{x}_{i,j}\|_2^2 - \mathcal{M}_c$, then the gradient back-propagation of the class-identity term can be formulated as follows:
\begin{equation}
\label{eq_12}
     c(\mathbf{X}_i, \mathbf{\Omega}) = \left\{
    \begin{array}{l}
        \frac{\partial \mathbb{C}(\mathbf{X}_i, \mathbf{\Omega})}{\partial \mathbf{\Omega}}, \hspace{0.1cm} if \hspace{0.1cm} \mathcal{C} > 0;\\
        \hspace{0.57cm}0\hspace{0.59cm}, \hspace{0.1cm}else.
    \end{array} \right.,
\end{equation}
where $\frac{\partial \mathbb{C}}{\partial \mathbf{\Omega}}$ can be formulated as follows:
\begin{equation}
\label{eq_13}
    \frac{\partial \mathbb{C}}{\partial \mathbf{\Omega}} = \frac{1}{\mathcal{Z}_c} \sum\limits_{j = 1}^M 2(\mathbf{c}_i - \mathbf{x}_{i,j})\cdot \frac{\partial{\mathbf{c}}_i - \partial \mathbf{x}_{i,j}}{\partial \mathbf{\Omega}}.
\end{equation}

The gradient back-propagation of the proposed relative distance term is consisted of two parts: $s = t(\mathbf{X}_i, \mathbf{\Omega}) + \lambda p(\mathbf{X}_i, \mathbf{\Omega})$, where $t(\cdot)$ denotes the gradient of the triplet distance term and $p(\cdot)$ represents the gradient of the marginal pairwise distance term. Similarly, we define $\mathcal{T} = \mathcal{M}_t + \mathrm{T}(\mathbf{x}_{i,l}^A,\mathbf{x}_{j,s}^B, \mathbf{x}_{k,r}^B)$, therefore the gradient of the triplet term can be formulated as follows:
\begin{equation}
\label{eq_14}
     t(\mathbf{X}_i, \mathbf{\Omega}) = \left\{
    \begin{array}{l}
        \frac{\partial \mathbb{T}(\mathbf{X}_i, \mathbf{\Omega})}{\partial \mathbf{\Omega}}, \hspace{0.1cm} if \hspace{0.1cm} \mathcal{T} > 0;\\
        \hspace{0.57cm}0\hspace{0.57cm}, \hspace{0.1cm}else.
    \end{array} \right.,
\end{equation}
where $\frac{\partial \mathbb{T}}{\partial \mathbf{\Omega}}$ can be formulated as follows:
\begin{equation}
\label{eq_15}
\begin{aligned}
    \frac{\partial \mathbb{T}}{\partial \mathbf{\Omega}} =  \frac{1}{\mathcal{Z}_t}\hspace{-0.1cm} \sum\limits_{j,k = 1}^N\sum\limits_{l,r,s = 1}^M \hspace{-0.1cm} 2\mathrm{P}_{i,j}^{l,s}(\mathbf{x}_{i,l}^A - \mathbf{x}_{j,s}^B)\cdot \frac{\partial \mathbf{x}_{i,l}^A - \partial \mathbf{x}_{j,s}^B}{\partial \mathbf{\Omega}}\\
    - 2\mu \mathrm{N}_{i,k}^{l,r}(\mathbf{x}_{i,l}^A - \mathbf{x}_{k,r}^B)\cdot \frac{\partial \mathbf{x}_{i,l}^A - \partial \mathbf{x}_{k,r}^B}{\partial \mathbf{\Omega}}\\
    - 2\nu \mathrm{N}_{i,k}^{l,r}(\mathbf{x}_{j,s}^B - \mathbf{x}_{k,r}^B)\cdot \frac{\partial \mathbf{x}_{j,s}^B - \partial \mathbf{x}_{k,r}^B}{\partial \mathbf{\Omega}}
\end{aligned}.
\end{equation}

By defining $\mathcal{P} = \mathcal{C}_i - \mathrm{G}_{i,j}^{l,s}(\mathcal{M}_p - \|\mathbf{x}_{i,l}^A - \mathbf{x}_{j,s}^B\|_2^2)$, then the gradient back-propagation of the pairwise term can be formulated as follows:
\begin{equation}
\label{eq_16}
    p(\mathbf{X}_i, \mathbf{\Omega}) = \left\{
    \begin{array}{l}
        \frac{\partial \mathbb{P}(\mathbf{X}_i, \mathbf{\Omega})}{\partial \mathbf{\Omega}}, \hspace{0.1cm} if \hspace{0.1cm} \mathcal{P} > 0;\\
        \hspace{0.67cm}0\hspace{0.51cm}, \hspace{0.1cm}else.
    \end{array} \right.,
\end{equation}
where $\frac{\partial \mathbb{P}}{\partial \mathbf{\Omega}}$ is defined as follows:
\begin{equation}
\label{eq_17}
    \frac{\partial \mathbb{P}}{\partial \mathbf{\Omega}} = \frac{1}{\mathcal{Z}_p} \sum\limits_{j = 1}^N \sum\limits_{l,s = 1}^M 2 \mathrm{G}_{i,j}^{l,s} (\mathbf{x}_{i,l}^A - \mathbf{x}_{j,s}^B) \cdot \frac{\partial \mathbf{x}_{i,l}^A - \partial \mathbf{x}_{j,s}^B}{\partial \mathbf{\Omega}}.
\end{equation}

\begin{algorithm}[tb]
   \caption{The S2S gradient descent algorithm}
   \label{alg}
\begin{algorithmic}
   \STATE {\bfseries Input:} \\
   \hspace{0.5 cm} Training samples $\mathbf{X}$, learning rate $\omega$, maximum iterations $H$, initialization to weight parameters $\mu$ and $\nu$, updating rate $\eta$, margin parameters $\mathcal{M}_c, \mathcal{M}_t, \mathcal{C}_p$ and $\mathcal{M}_p$, and weight parameters $\alpha, \beta$ and $\lambda$.
   \STATE {\bfseries Output:} \\
   \hspace{0.5 cm}The network parameters $\mathbf{\Omega}$.
   \REPEAT
   \STATE 1. Calculate the output feature representations of $\mathbf{x}_{i,j}$, $\mathbf{x}_{i,l}^A$, $\mathbf{x}_{j,s}^B$ and $\mathbf{x}_{k,r}^B$ used in the class-identity term and the relative distance term in a mini-batch by the forward propagation.\\
   \REPEAT
   \STATE a) Update the weight parameters $\mu$ and $\nu$ according to Eq.~\eqref{eq_18}, Eq.~\eqref{eq_19} and Eq.~\eqref{eq_20};
   \STATE b) Calculate $\frac{\partial \mathbb{C}}{\partial \mathbf{\Omega }}$, $\frac{\partial \mathbb{T}}{\partial \mathbf{\Omega }}$ and $\frac{\partial \mathbb{P}}{\partial \mathbf{\Omega }}$ according to Eq.~\eqref{eq_13}, Eq.~\eqref{eq_15} and Eq.~\eqref{eq_17}, respectively;
   \STATE c) Increment the gradient $\frac{\partial \mathrm{L}}{\partial \mathbf{\Omega }}$ according to Eq.~\eqref{eq_11}, Eq.~\eqref{eq_12}, Eq.~\eqref{eq_14} and Eq.~\eqref{eq_16}, respectively;
   \UNTIL{Traverse all the set units in the training samples in each mini-batch.}
   \STATE 2. Update $\mathbf{\Omega}_{h+1} = \mathbf{\Omega}_h - {\tau_h}\frac{\partial \mathrm{L}}{\partial \mathbf{\Omega}_h}$ and $h \leftarrow h + 1$.
   \UNTIL{$h > H$}
\end{algorithmic}
\end{algorithm}

As described above, the weights $\mu$ and $\nu$ can be adaptively learned in the training process, we update them by using the gradient back-propagation method. In order to simplify the problem, we define $\mu = \psi + \varphi, \nu = \psi - \varphi$, therefore they can be updated by only updating $\varphi$. The partial derivative of the triplet distance term with respect to $\varphi$ can be formulated as follows:
\begin{equation}
\label{eq_18}
     r(\mathbf{X}_i, \mathbf{\Omega}) = \left\{
    \begin{array}{l}
        \frac{\partial \mathrm{T}(\mathbf{X}_i, \mathbf{\Omega})}{\partial \varphi}, \hspace{0.1cm} if \hspace{0.1cm} \mathcal{T} > \mathcal{M}_t;\\
        \hspace{0.57cm}0\hspace{0.57cm}, \hspace{0.1cm}else.
    \end{array} \right.,
\end{equation}
where $\frac{\partial \mathrm{T}}{\partial \varphi}$ is computed as follows:
\begin{equation}
\label{eq_19}
    \frac{\partial \mathrm{T}}{\partial \varphi} = 2\mathrm{N}_{i,k}^{l,r}(\|\mathbf{x}_{i,l}^A - \mathbf{x}_{j,s}^B\|_2^2 - \|\mathbf{x}_{j,s}^B - \mathbf{x}_{k,r}^B\|_2^2).
\end{equation}

We adopt the momentum method~\cite{He_Zhang_Ren:2015} to update $\varphi$, which is formulated as follows:
\begin{equation}
\label{eq_20}
    \varphi = \varphi - \eta \cdot r,
\end{equation}
where $\eta$ is the updating rate. It can be clearly seen that when $\|\mathbf{x}_{i,l}^A - \mathbf{x}_{j,s}^B\|_2^2 > \|\mathbf{x}_{j,s}^B - \mathbf{x}_{k,r}^B\|_2^2$, namely $r < 0$, then $\mu$ will be decreased while $ \nu$ will be increased; and vice verse. As a result, the strength of back-propagation to each sample in the same triplet unit will be adaptively tuned, and accordingly the anchor and the positive will be clustered, and the negative one will be far away from the hyper-line expanded by the anchor and the positive.

\begin{figure}[!htb]
\centering
    \includegraphics[height = 6.5cm, width = 7.2cm]{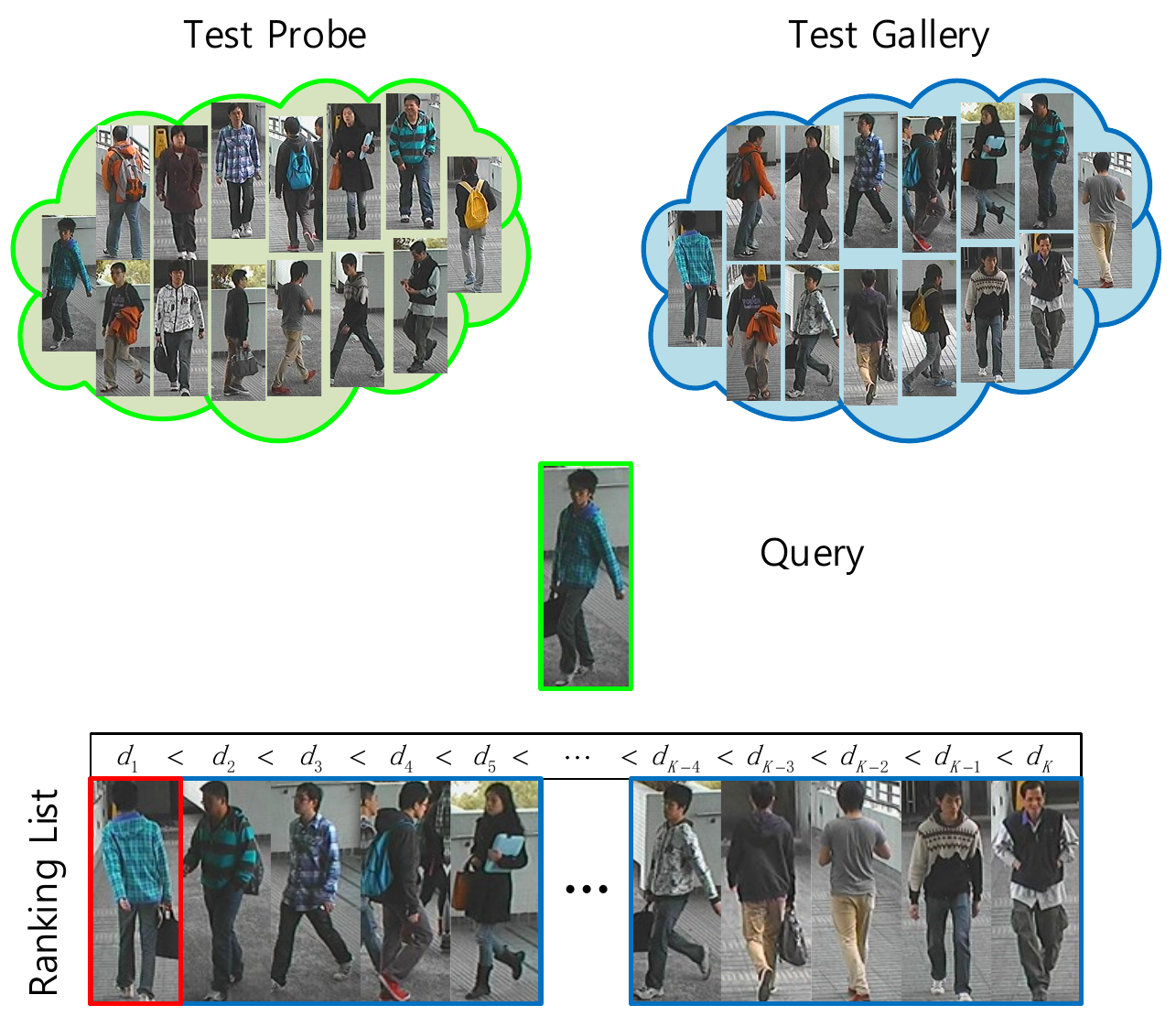}\\
    \caption{The illustration of how to evaluate our method. For each query image from the test probe set, we compute the pairwise distance with all the candidates in the test gallery set. The distances are then ranked to find out which top $n$ can find the corrected matches.}
    \label{fig_8}
\end{figure}

From the above derivations, it is clear that the gradients can be easily calculated given the values of $\mathbf{x}_{i,j}, \mathbf{x}_{i,l}^A, \mathbf{x}_{j,s}^B, \mathbf{x}_{k,r}^B$ and ${\partial{\mathbf{x}_i^j}}/{\partial \mathbf{\Omega}}, {\partial{\mathbf{x}_{i,l}^A}}/{\partial \mathbf{\Omega}}, {\partial{\mathbf{x}_{j,s}^B}}/{\partial \mathbf{\Omega}}, {\partial{\mathbf{x}_{k,r}^B}}/{\partial \mathbf{\Omega}}$ in each mini-batch, in which they can be obtained by separately running the forward and backward propagation for each image in both the pairwise and triplet units. As the algorithm needs to go through all the images sets to accumulate the gradients in each iteration, we call it the S2S gradient descent algorithm. We show the overall process in Algorithm~\ref{alg}.

\section{Experiment}
\label{sec_exp}

\subsection{Dataset and Setup}
We evaluated our method on four widely used benchmark datasets, specifically the CUHK01~\cite{Li_Wang:2013}, the CUHK03~\cite{Li_Zhao_Xiao:2014}, the PRID2011~\cite{Hirzer_Beleznai_Roth:2011} and the Market1501~\cite{Zheng_Shen_Tian:2015} datasets. In these datasets, each person has at least two images under each camera view.

{\bf CUHK01:} The dataset contains 971 persons captured from two camera views in a campus environment, and there are two images for each person under every camera view. We utilized the same protocol as~\cite{Wang_Zuo_Lin:2016}, where 871 person images are used for training and the rest for testing.

{\bf CUHK03:} There are 13164 images from 1360 persons in the CUHK03 dataset. All the person images are captured from six cameras, and each person only has two camera views. We followed the protocol in~\cite{Ahmed_Jones_Marks:2015} for training/testing partition.

{\bf PRID2011:} The dataset contains video clips of 749 persons from two disjoint cameras, each clip with 5 to 675 frames. Following the protocol in~\cite{You_Wu_Li:2016}, we only considered the first 200 persons who appear in both cameras.

{\bf Market1501:} The dataset contains 32668 images of 1501 persons. Each person is captured by at least two cameras and six at most. We used the provided training/testing partition, under both the single-query and multi-query evaluation settings as in~\cite{Zhang_Xiang_Gong:2016}.

\begin{table}[h]
\footnotesize
\caption{Matching rates(\%) on the CUHK01 dataset.}
\begin{center}
\label{tab_1}
\begin{tabular}{ c| c | c | c| c | c}
\hline
Methods & Top1 & Top5 & Top10 & Top15 & Top20\\
\hline
\hline
KISSME~\cite{Koestinger_Hirzer_Wohlhart:2012}           & 29.40 & 59.34 & 71.45 & 80.09 & 88.12 \\
LMNN~\cite{Weinberger_Blitzer_Saul:2005}                & 21.17 & 49.49 & 61.12 & 69.93 & 78.32 \\
IDLA~\cite{Ahmed_Jones_Marks:2015}                      & 65.00 & 89.33 & 92.04 & 93.74 & 96.51 \\
JSC~\cite{Wang_Zuo_Lin:2016}                            & 65.71 & 89.41 & 92.52 & 93.74 & 96.63 \\
CDVM~\cite{Lin_Wang_Zuo:2016}                           & 66.50 & \textcolor{blue}{93.00} & \textcolor{blue}{96.50}
                                                        & \textcolor{red}{\bf{99.00}} & \textcolor{red}{\bf{99.00}}\\
\hline
Our Method (P2P)                                        &\textcolor{blue}{70.10} & 85.82 & 95.43
                                                        &96.35 &97.20 \\
Our Method (S2S)                                        &\textcolor{red}{\bf{77.89}}  &\textcolor{red}{\bf{93.22}}  &\textcolor{red}{\bf{96.61}}
                                                        &\textcolor{blue}{98.23}  &\textcolor{blue}{98.68}  \\
\hline
\end{tabular}
\end{center}
\end{table}

{\bf Parameter setting}\footnote{The parameters used in this paper are obtained based on the performance on the CUHK01 dataset, one can achieve better results by choosing more suitable parameters using the cross validation method.} The weights were initialized from two zero-mean Gaussian distribution with the standard deviations from $0.01$ to $0.001$, respectively. The bias terms were set to $0.0$. The learning rate $\omega = 0.01$, the updating rate $\eta = 0.001$, the weight parameters $\alpha = 0.1, \beta = 0.01, \lambda = 0.15$, the direction control parameters $\mu = 0.6, \nu = 0.4$ and the margin parameters $\mathcal{C}_p = 0.175, \mathcal{M}_p = 0.325, \mathcal{M}_t = 1.0$.

{\bf Protocol} The dataset was separated into a training set and a testing set, and images of a same person can only appear in either set. The testing set was further divided into a probe set and a gallery set, and the two sets contained images of the same person from different views. The result was evaluated by cumulatively matching characteristic~(CMC) curve~\cite{Gray_Brennan_Tao:2007} which is an estimation of finding the corrected math in the top $n$ match, as shown in Fig.~\ref{fig_8}. The final performance was averaged over ten experiment attempts.

\subsection{Results}

{\bf Comparison Results} We compared the results of our method with several state-of-the-art works on the four benchmark datasets, specifically KISSME~\cite{Koestinger_Hirzer_Wohlhart:2012}, LADF~\cite{Li_Chang_Liang:2013}, LF~\cite{Pedagadi_Orwell_Velastin:2013}, ME~\cite{Paisitkriangkrai_Shen:2015}, kLFDA~\cite{Xiong_Gou_Camps:2014}, SCSP~\cite{Chen_Yuan_Chen:2016}, CDVM~\cite{Lin_Wang_Zuo:2016}, LMNN~\cite{Weinberger_Blitzer_Saul:2005}, IDLA~\cite{Ahmed_Jones_Marks:2015}, FPNN~\cite{Li_Zhao_Xiao:2014}, LOMO+XQDA~\cite{Liao_Hu_Zhu:2015}, and LSSCDL~\cite{Zhang_Li_Lu:2016}. In order to analyze how each ingredient contributes to the final performance improvement, we report the results of our method from two aspects: 1) To evaluate the performance of our method with the proposed point to point~(P2P) constraint, we got rid of the class-identity term and the pairwise term in our method, therefore only the P2P information remained in our method; 2) To reveal how the set to set~(S2S) constraint contributes to the performance improvement, we report the final performance by jointly using the class-identity term, the triplet term and the pairwise term in our method. We report detailed comparison results on the four datasets in Table~\ref{tab_1} to Table~\ref{tab_4}. In all tables, the second best performance is highlighted in blue, and the best performance is in red.

For the CUHK01 dataset, we compared our methods with both the traditional methods and the deep learning based methods, and the results are shown in Table~\ref{tab_1}. We can see from the results that, our two methods outperformed the deep learning based methods, such as IDLA~\cite{Ahmed_Jones_Marks:2015}, JSC~\cite{Wang_Zuo_Lin:2016} and CDVM~\cite{Lin_Wang_Zuo:2016}. In particular, our two methods outperformed the CDVM method by $3.60\%$ and $11.39\%$ in Top 1 accuracy, respectively. In addition, by considering the S2S information, the S2S method outperformed the P2P method by $7.79\%$ in Top 1 accuracy .

\begin{table}[h]
\footnotesize
\caption{Matching rates(\%) on the CUHK03 dataset.}
\begin{center}
\label{tab_2}
\begin{tabular}{ c| c | c | c| c | c}
\hline
Methods & Top1 & Top5 & Top10 & Top15 & Top20\\
\hline
\hline
FPNN~\cite{Li_Zhao_Xiao:2014}                 & 20.65 & 51.02 & 68.83 & 76.38 & 81.45 \\
LOMO+XQDA~\cite{Liao_Hu_Zhu:2015}             & 52.20 & 81.29 & 90.94 & 94.21 & 95.01 \\
IDLA~\cite{Ahmed_Jones_Marks:2015}            & 54.74 & 87.59 & 94.01 & 95.02 & 95.41\\
LSSCDL~\cite{Zhang_Li_Lu:2016}                & 57.00 & 84.38 & 90.93 & 94.32 & 95.12\\
CDVM~\cite{Lin_Wang_Zuo:2016}                 & \textcolor{blue}{58.39} & 85.56 & 92.57 & 94.48 & 96.60\\
\hline
Our Method (P2P)                              & 54.61 & \textcolor{blue}{86.80} & \textcolor{blue}{93.12}
                                              & \textcolor{blue}{96.22} & \textcolor{blue}{98.23} \\
Our Method (S2S)                              & \textcolor{red}{\bf{63.58}}  & \textcolor{red}{\bf{89.17}} & \textcolor{red}{\bf{93.75}}
                                              & \textcolor{red}{\bf{96.35}}  & \textcolor{red}{\bf{98.25}} \\
\hline
\end{tabular}
\end{center}
\end{table}

\begin{table}[h]
\footnotesize
\caption{Matching rates(\%) on the PRID2011 dataset.}
\begin{center}
\label{tab_3}
\begin{tabular}{ c| c | c | c | c}
\hline
Methods & Top1 & Top5 & Top10  & Top20\\
\hline
\hline
KISSME~\cite{Koestinger_Hirzer_Wohlhart:2012}           & 28.54 & 59.78 & 72.13 & 83.26 \\
LF~\cite{Pedagadi_Orwell_Velastin:2013}                 & 26.40 & 56.07 & 69.89 & 81.12 \\
LMNN~\cite{Weinberger_Blitzer_Saul:2005}                & 14.38 & 38.09 & 50.22 & 67.19 \\
LADF~\cite{Li_Chang_Liang:2013}                         & 8.20  & 20.45 & 29.89 & 42.25 \\
TDL~\cite{You_Wu_Li:2016}                               & 30.22 & 59.10 & 74.04 & 88.43 \\
\hline
Our Method (P2P)                                        &\textcolor{blue}{69.41} &\textcolor{red}{\bf{95.63}} &\textcolor{blue}{100.00}
                                                        &\textcolor{blue}{100.00}\\
Our Method (S2S)                                        &\textcolor{red}{\bf{72.54}}  &\textcolor{blue}{94.61}  &\textcolor{red}{\bf{100.00}}
                                                        &\textcolor{red}{\bf{100.00}}\\
\hline
\end{tabular}
\end{center}
\end{table}

\begin{table}[h]
\footnotesize
\caption{Matching rates(\%) on the Market1501 dataset.}
\begin{center}
\label{tab_4}
\begin{tabular}{ c| c | c || c | c  c}
\hline
\multicolumn{1}{c|}{\multirow{2}{*}{Methods}} &
\multicolumn{2}{c||}{Single-Query} &
\multicolumn{2}{c}{Multi-Query}\\
\cline{2-5}
& Top1 & mAP & Top1  & mAP\\
\hline
\hline
Bow~\cite{Zheng_Shen_Tian:2015}                         & 34.38  & 14.10 & 42.64 & 19.47\\
kLFDA~\cite{Xiong_Gou_Camps:2014}                       & 51.37  & 24.43 & 52.67 & 27.36\\
KISSME~\cite{Koestinger_Hirzer_Wohlhart:2012}           & 40.50  & 19.02 & $--$  & $--$ \\
LDNS~\cite{Zhang_Xiang_Gong:2016}                       &61.02   & 35.68 & 71.56 & 46.03 \\
SCSP~\cite{Chen_Yuan_Chen:2016}                         & 51.90  & 26.35 & $--$  & $--$\\
\hline
Our Method (P2P)                                        &\textcolor{blue}{62.41} & \textcolor{blue}{36.21}
                                                        &\textcolor{blue}{77.19} & \textcolor{blue}{49.24}\\
Our Method (S2S)                                        &\textcolor{red}{\bf{65.32}}  & \textcolor{red}{\bf{39.83}}
                                                        &\textcolor{red}{\bf{80.49}}  & \textcolor{red}{\bf{52.69}}\\
\hline
\end{tabular}
\end{center}
\end{table}

In Table~\ref{tab_2}, we report the final comparison results with the state-of-the-art methods on the CUHK03 dataset. Our S2S method achieved the best performance in all comparison groups from Top 1 to Top 20. Compared with the previous best method CDVM~\cite{Lin_Wang_Zuo:2016}, our P2P method scored $3.78\%$ lower in Top 1 accuracy, while our S2S method outperformed it by $5.19\%$, which again shows the benefit from the S2S information that gives $8.97\%$ improvement over the P2P metric in Top 1 accuracy.

The PRID2011 dataset is specially designed for video based person Re-ID, however we do not use any video information in our S2S method, therefore we could compare our method with the state-of-the-art methods in a multi-shot setting~\cite{You_Wu_Li:2016}. The comparison results are shown in Table~\ref{tab_3}, and our P2P method won the second best performance and our S2S method achieved the best performance in all categories from Top 1 to Top 20. Compared with the previous best method TDL~\cite{You_Wu_Li:2016}, the two proposed methods outperformed it significantly by $39.19\%$ and $42.32\%$ in Top 1 accuracy, respectively. In addition, the S2S method won the P2P method by $3.13\%$ in Top 1 accuracy by incorporating the S2S information into the relative distance term.

The Market1501 dataset is a new and large scale dataset for person Re-ID. The best existing performance is achieved by the conventional method. We show comparison results in Table~\ref{tab_4}, where we evaluated the performance using both the CMC curve and the mAP~\cite{Zheng_Shen_Tian:2015} value under the single-query and multi-query evaluation settings, respectively. Compared with the best existing method LDNS~\cite{Zhang_Xiang_Gong:2016}, the two proposed methods outperformed it by $1.39\%$ and $4.30\%$ in Top 1 accuracy under the single-query setting, and $5.63\%$ and $8.93\%$ in Top 1 accuracy under the multi-query setting, respectively. In addition, the S2S method won the P2P method by $2.91\%$ and $3.30\%$ in Top 1 accuracy under both the single-query and multi-query evaluation settings, respectively. For the mAP evaluation, the second best performance was achieved by the P2P method as compared with the existing methods, while the S2S method further outperformed the P2P method under both single-query and multi-query settings by incorporating the S2S information.

\begin{figure*}[!htb]
\centering
    \begin{tabular}{ccc}
        \includegraphics[height = 4.5cm, width = 5.0cm]{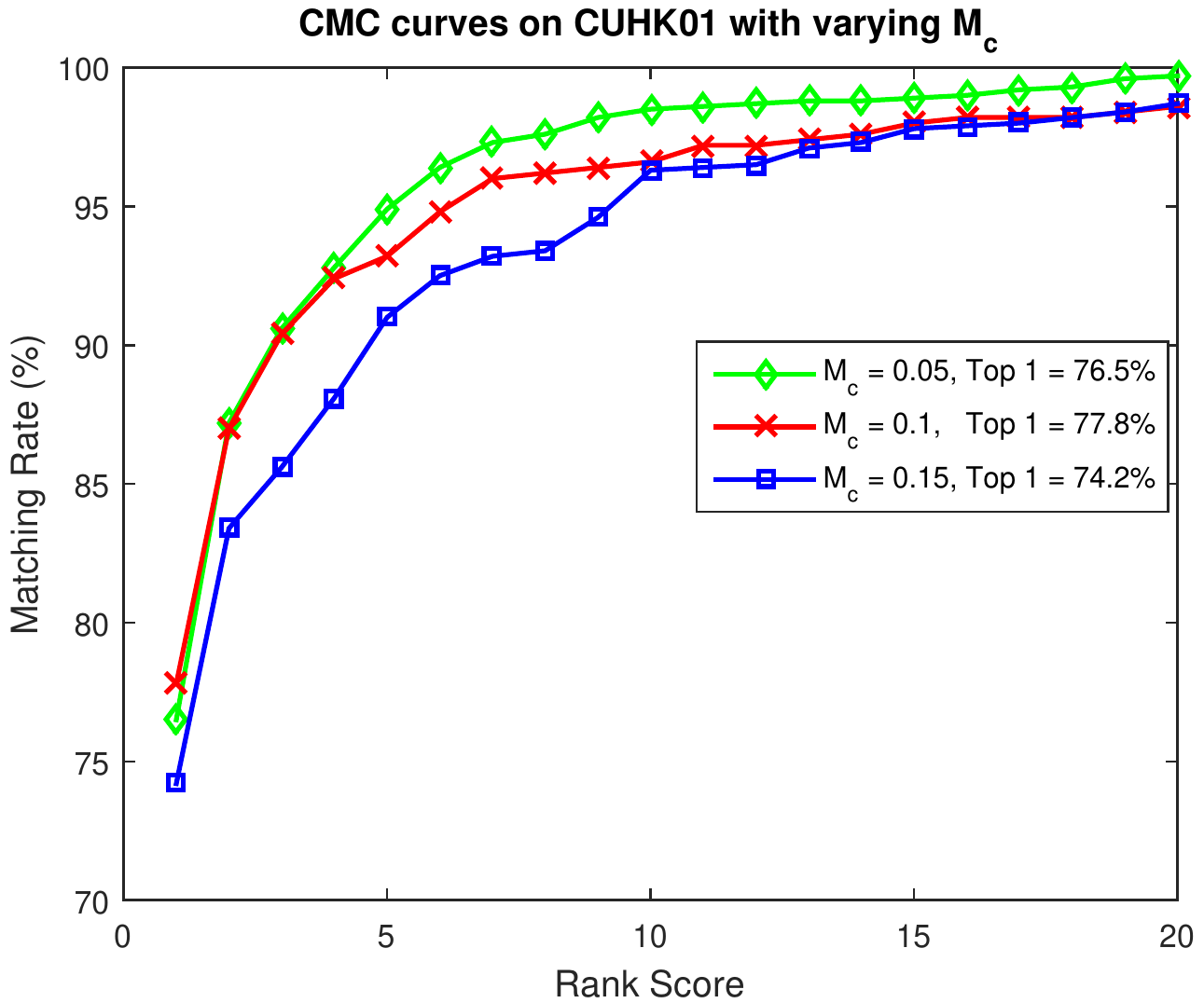}&
        \includegraphics[height = 4.5cm, width = 5.0cm]{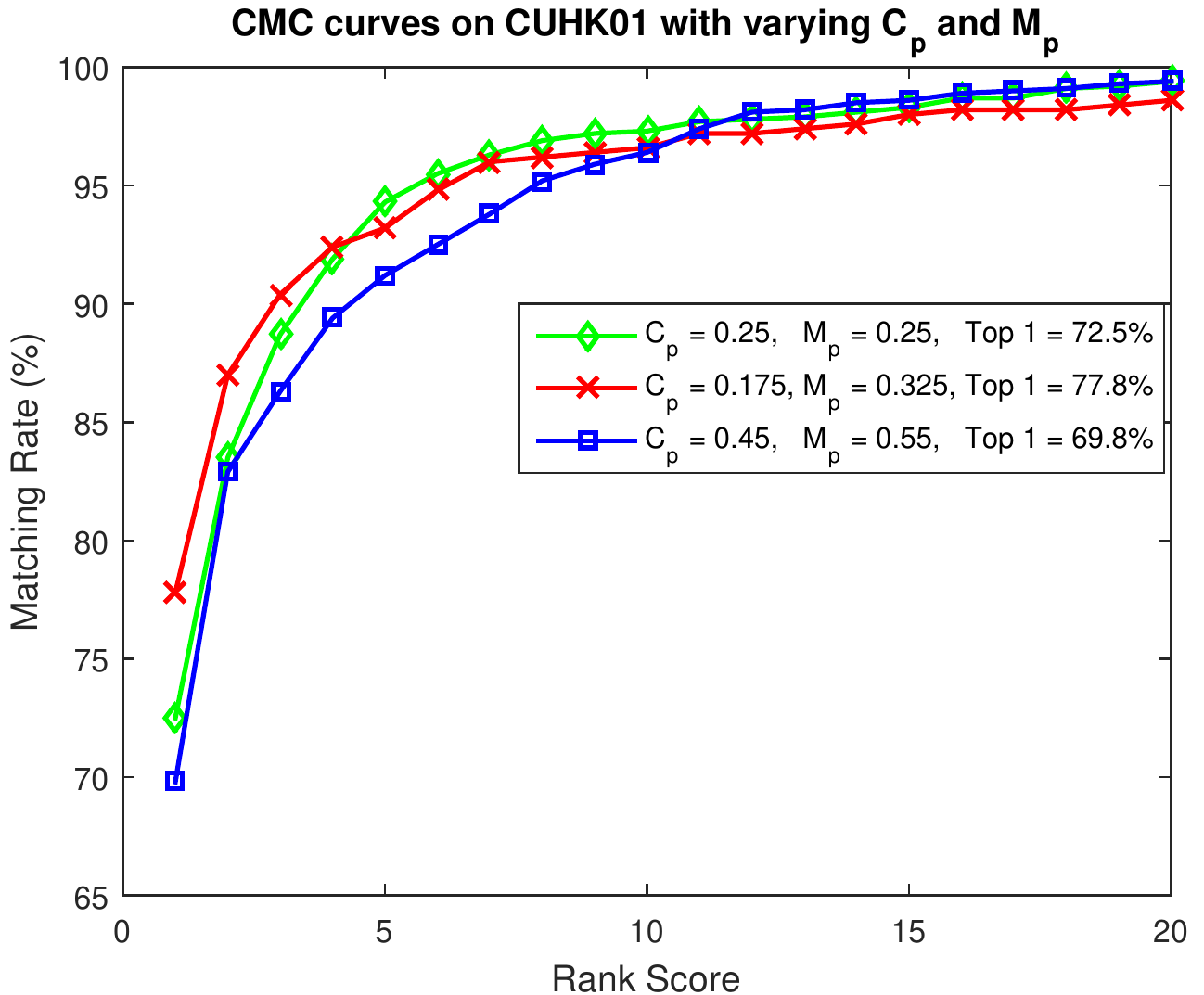}&
        \includegraphics[height = 4.5cm, width = 5.0cm]{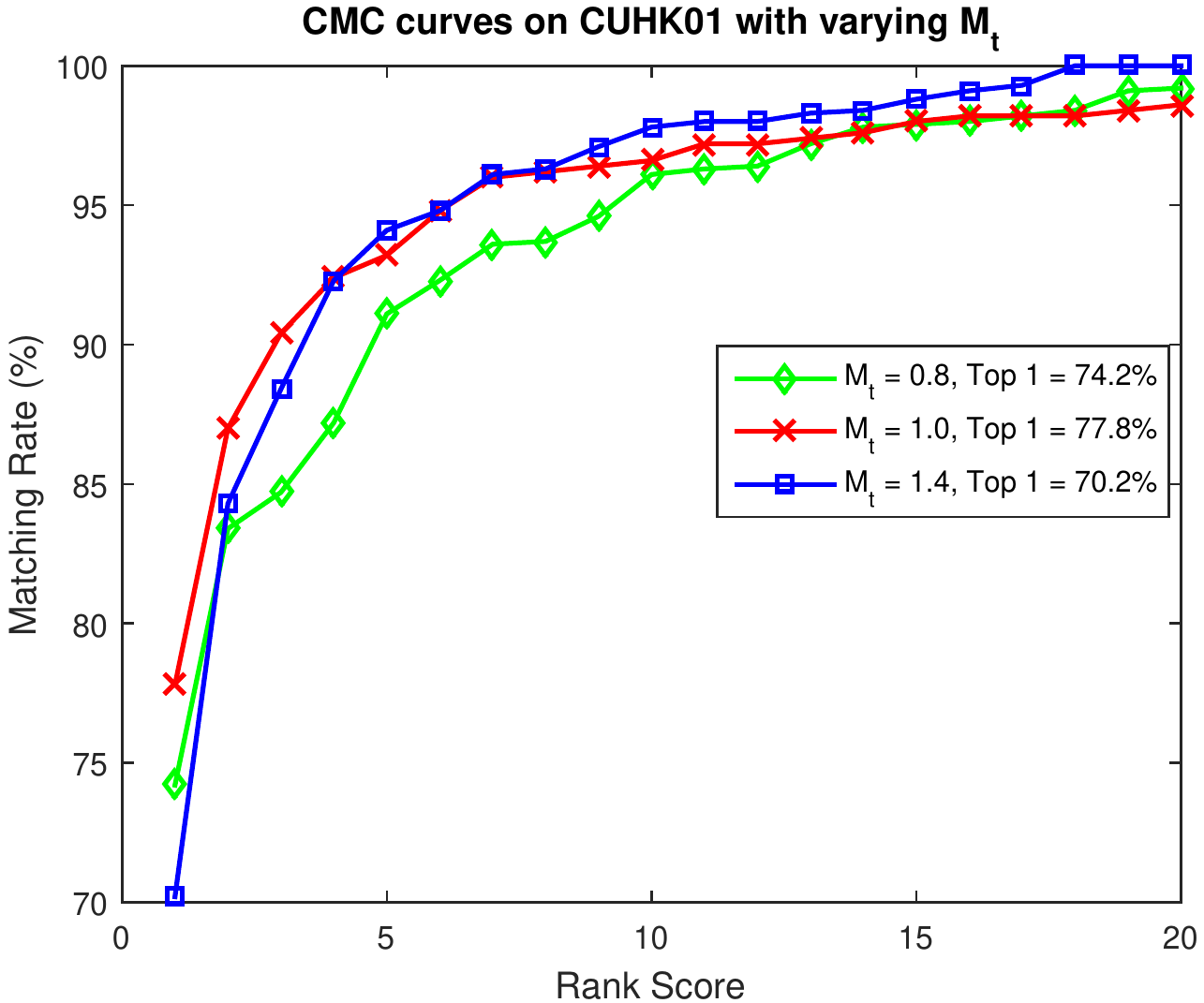}
    \end{tabular}
    \caption{CMC curves on the CUHK01 dataset with varying margin parameters, where (a) shows the matching results with varying $\mathcal{M}_c$ and setting $\mathcal{M}_t = 1.0$ amd $\mathcal{C}_p = 0.175, \mathcal{M}_p = 0.325$ (b) shows the matching results with varying $\mathcal{C}_p, \mathcal{M}_p$ and setting $\mathcal{M}_c=0.1, \mathcal{M}_t = 1.0$; and (c) shows the matching results with varying $\mathcal{M}_t$ and setting $\mathcal{M}_c = 0.1, \mathcal{C}_p = 0.175$ and $\mathcal{M}_p = 0.325$.}
    \label{fig_6}
\end{figure*}

\begin{figure}[!htb]
\centering
    \begin{tabular}{c}
       \includegraphics[height = 6.5cm, width = 8.5cm]{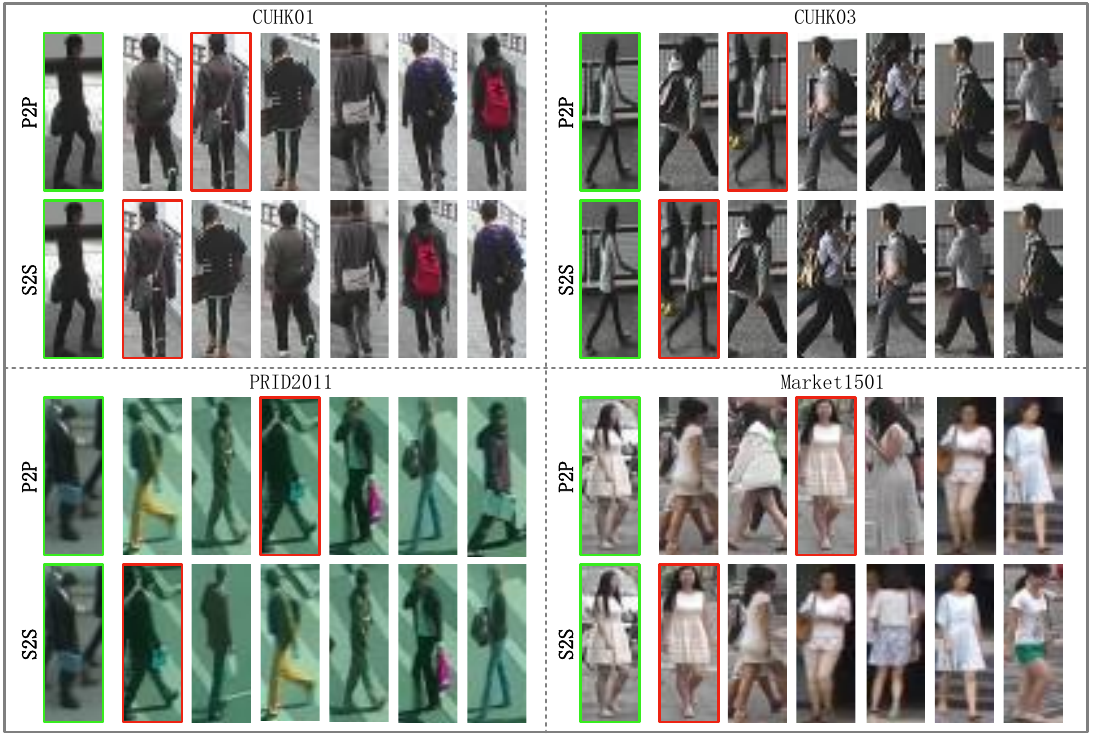}
    \end{tabular}
    \caption{The ranking results on the four benchmark datasets in the single-shot setting, where person in green rectangle denotes the query image and person in red rectangle represents the matched candidate.}
    \label{fig_5}
\end{figure}

{\bf Analysis} To obtain more insight between the P2P and the S2S methods, some typical ranking examples on the four benchmark datasets are shown in Fig.~\ref{fig_5}. Specifically, the green rectangle denotes the query image from the probe set and the red rectangle represents the matched candidate in the gallery set. For each dataset, we give the comparison ranking results of the two methods, i.e. the P2P method result and the S2S method, in the first and the second row respectively. According to the results, we can see that the S2S method is more robust than the P2P method in dealing with appearance variations caused by view angle, body pose, mutual occlusion and light condition. More importantly, the S2S method can effectively select the matched candidate out, as compared with the P2P method in the CUHK01 dataset, no matter how the view angle changes in the gallery set. On the other side, we find that none of the two methods can distinguish image details. For example, the P2P method ranks the person wearing a yellow pant before the correct candidate with a light blue bag in hand in the PRID2011 dataset, and the S2S method ranks the people wearing a green short before other more similar candidates in white dress in the Market1501 dataset. Considering the fact that our deep CNN does not perform salient information detection, none of the two methods can recognize such subtle differences of targets. In our future work, we will strive to find an optimal saliency detection modular in the framework of deep CNN, so as to boost the ranking performance by alleviating the influence of these details.

\begin{table}[h]
\footnotesize
\caption{Influence of the direction control parameters.}
\begin{center}
\label{tab_5}
\begin{tabular}{ c| c | c || c | c || c |c c}
\hline
\multicolumn{1}{c|}{\multirow{2}{*}{Datasets}} &
\multicolumn{2}{c||}{$\mu \hspace{-0.1cm} = \hspace{-0.1cm}1.0, \nu \hspace{-0.1cm} = \hspace{-0.1cm} 0.0$} &
\multicolumn{2}{c||}{$\mu \hspace{-0.1cm} = \hspace{-0.1cm}0.6, \nu \hspace{-0.1cm} = \hspace{-0.1cm} 0.4$} &
\multicolumn{2}{c}{$\mu \hspace{-0.1cm}  = \hspace{-0.1cm}0.4, \nu \hspace{-0.1cm} = \hspace{-0.1cm} 0.6$}\\
\cline{2-7}
&Top1 & Top5 & Top1  & Top5 & Top1  & Top5\\
\hline
\hline
CUHK01            &69.32 & 91.43 & \textcolor{red}{\bf{77.89}} & \textcolor{red}{\bf{93.22}} & \textcolor{blue}{74.85} & \textcolor{blue}{92.31}\\
CUHK03            &57.32 & 87.31 & \textcolor{red}{\bf{63.58}} & \textcolor{red}{\bf{89.17}} & \textcolor{blue}{60.23} & \textcolor{blue}{88.54}\\
PRID2011          &64.25 & 92.08 & \textcolor{red}{\bf{72.54}} & \textcolor{red}{\bf{94.61}} & \textcolor{blue}{69.38} & \textcolor{blue}{92.15}\\
Market1501        &57.42 & 78.54 & \textcolor{red}{\bf{65.32}} & \textcolor{red}{\bf{85.45}} & \textcolor{blue}{63.24} & \textcolor{blue}{82.21}\\
\hline
\end{tabular}
\end{center}
\end{table}
\subsection{Influence of Parameters}
Emperically, the margin parameters $\mathcal{M}_c, \mathcal{M}_t, \mathcal{C}_p, \mathcal{M}_p$, the weight parameters $\alpha, \lambda$ and the direction control parameters $\mu, \nu$ have major effects to the final ranking performance of our method. In the following paragraphs, we give experimental analysis of our method on the CUHK01 dataset by changing one parameter at a time to study its influence.

\begin{figure}[!htb]
\centering
    \begin{tabular}{cc}
        \includegraphics[height = 4.2cm, width = 4.0cm]{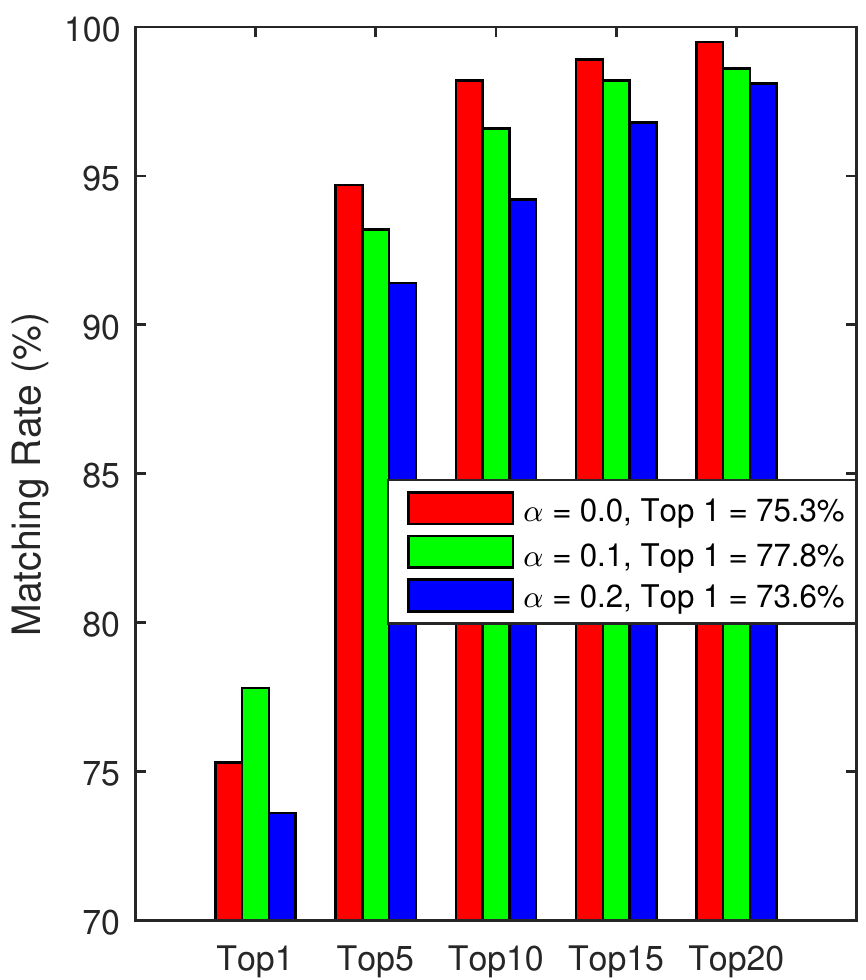}&
        \includegraphics[height = 4.2cm, width = 4.0cm]{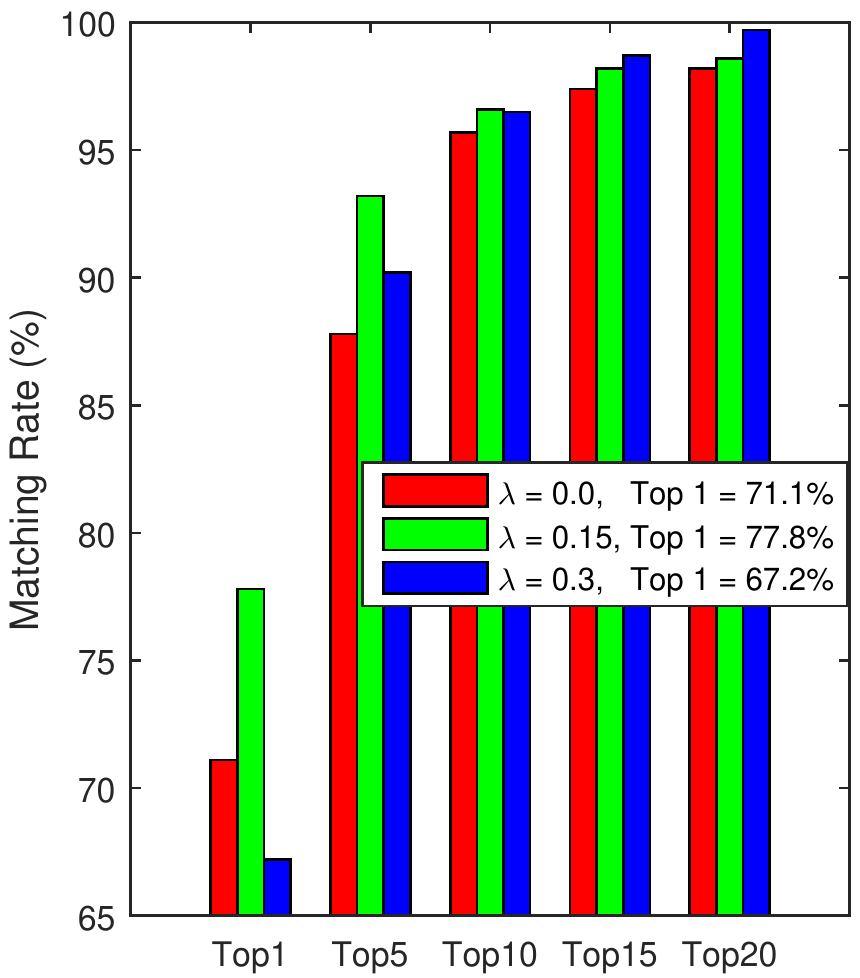}
    \end{tabular}
    \caption{CMC curves on the CUHK01 dataset with varying weight parameters, where (a) shows the matching results with varying $\alpha$ and setting $\lambda = 0.15$, and (b) shows the matching results with varying $\lambda$ and setting $\alpha = 0.1$.}
    \label{fig_7}
\end{figure}

The influence of margin parameters $\mathcal{M}_c, \mathcal{M}_t, \mathcal{C}_p, \mathcal{M}_p$ is shown in Fig.~\ref{fig_6}, and our method achieved the best performance by setting $\mathcal{M}_c = 0.1, \mathcal{C}_p = 0.175, \mathcal{M}_p = 0.325$ and $\mathcal{M}_t = 1.0$. We can derive the following three conclusions: 1) Parameter $\mathcal{M}_c$ is not significant, because the intra-class difference under each camera view is very small. According to the experience, we could set $\mathcal{M}_c = 0.1$ in all the experiments. 2) For parameters $\mathcal{C}_p, \mathcal{M}_p$, small down-margin $\mathcal{M}_p - \mathcal{C}_p$ will lead to over-fitting problem, and large upper-margin $\mathcal{M}_p + \mathcal{C}_p$ will cause numerical stability issue. 3) Similarly, large $\mathcal{M}_t$ will also lead to numerical stability problem, and small $\mathcal{M}_t$ will make  positive and negative candidates less distinguishable in the distance space.

For the weight parameters $\alpha$ and $\lambda$, their influence to the final results is shown in Fig.~\ref{fig_7}, and our method achieved its best performance by setting $\alpha = 0.1$ and $\lambda = 0.15$. From the results, we can see that $\alpha$ has little influence to the final ranking performance. Our analysis show that since the intra-class difference of individuals under the same camera view is significant smaller than that of the inter-class, they can get clustered together even with a small $\alpha$. On the contrary, $\lambda$ is a more sensitive parameter, as greater value may lead to  over-fitting problem, while smaller value may cause set information missing in the method.

Different from the conventional triplet formulation~\cite{Ding_Lin_Wang:2015} used in person Re-ID, our symmetric triplet framework introduces a weighed negative distance term to optimize the back-propagation pattern of each sample in one triplet unit. Therefore, the conventional triplet formulation is a special case of our symmetric triplet formulation by setting $\mu  = 1.0, \nu  = 0.0$ and $\eta = 0.0$ in our method. The comparison results reported in Table~\ref{tab_5} show that, our symmetric triplet framework outperforms the conventional one by $8.57\%, 6.26\%, 8.29\%$ and $7.90\%$ in Top 1 accuracy on the four datasets, respectively. Benefit from the weight updating strategy, the initial values of $\mu$ and $\nu$ have subtle impact to the final performance. We can see that the accuracy only falls by $3.04\%, 3.35\%, 3.16\%$ and $2.08\%$ when setting $\mu  = 0.4, \nu  = 0.6$ on the four datasets, respectively.

\section{Conclusion}
\label{sec_con}
In this paper, we propose a novel person re-identification method by set to set~(S2S) similarity comparison in the deep framework to perform integrated feature learning and fusion. The deep neural network learns the global features, local features and fused features in the global sub-network, local sub-network and fusion sub-network, respectively. The S2S distance metric can jointly keep the compactness of the intra-class samples under each camera view and maximize the relative distance between the intra-class set and inter-class set across different camera views, so as to back-propagate the gradients to optimize the parameters of the deep CNN. As a result, the learned deep ranking model can effectively distinguish different persons by learning the discriminative and robust feature representations. Extensive experimental results on the benchmark datasets, including the CUHK01, CUHK03, PRID2011 and Market1501, show that our method outperforms most of the state-of-the-art approaches in person re-identification.


\ifCLASSOPTIONcaptionsoff
  \newpage
\fi

{
\bibliographystyle{IEEEtran}
\bibliography{S2S_V7}
}

\end{document}